\documentclass[11pt]{article}

\usepackage{natbib}
\usepackage{amsmath,amssymb}
\usepackage{url}
\usepackage{arxiv}

\usepackage[font=footnotesize,labelfont=bf]{caption}
 \usepackage{graphicx}
 \usepackage{wrapfig}
 \usepackage{subcaption}

\title{What do Language Models Model? Transformers, automata, and the format of thought.} %
\shorttitle{What do LLMs model?}
\author{Colin Klein\\
  School of Philosophy\\
 The Australian National University\\
  \texttt{colin.klein@anu.edu.au} \\
 }
  \date{}

 \setnoticestring{\emph{Preprint of work in progress. Draft D5e1, compiled \today}}

\begin{document}

\maketitle

\begin{abstract}

What do large language models actually model? Do they tell us something about human capacities, or are they models of the corpus we've trained them on?  I give a non-deflationary defence of the latter position. Cognitive science tells us that  linguistic capabilities in humans rely \emph{supralinear} formats for computation. The transformer architecture, by contrast,  supports at best a \emph{linear} formats for processing. This argument will rely primarily on certain invariants of the computational architecture of transformers. I then suggest a positive story about what transformers are doing, focusing  on \citet{LiuTransformers22}'s intriguing speculations about shortcut automata.  I conclude with why I don't think this is a terribly deflationary story. Language is not (just) a means for expressing inner state but also a kind of `discourse machine' that lets us make new language given appropriate context. We have learned to use this technology in one way;  LLMs have also learned to use it too, but via very different means. 

\emph{Note on 26 August 2025:  this was a draft in progress for a long time. I'm not entirely happy with it, though I stand by the general argument. I've also started to carve chunks off for other projects. So I am archiving it for posterity, in case references to it crop up. If you'd like to discuss---and especially if you have thoughts about what I call substring invariance---please do get in touch. The overall project is active, it's just that this version is giving me a hard time.} 

\end{abstract}

\section{Introduction}

Large Language Models (LLMs) use language in ways that are strikingly similar the ways humans use language. Why? Presumably their training gives them a model of the world, and that model explains their performance.  But what do large language models actually model? 


The literature suggests two kinds of answer.  On the one hand, Large Language Models might model \emph{linguistic capacity}: That is, the performance of LLMs  is explained by the capacities that \emph{we} used when we made their training corpus. Much of the philosophical debate on this branch concerns which capacities LLMs have, and whether they properly have human capacities or merely simulate them: so, for example,  whether LLMs have grounded symbols \citep{PavlickSymbols23,ChalmersDoes23}, or have instrumental knowledge \citep{YildirimFrom24}, or have a theory of mind \citep{StrachanTesting24}, or are conscious \citep{ChalmersCould23, ButlinConsciousness23}, or whatever.  Much work in so-called \emph{mechanistic interpretability} \citep{OlahZoom20} similarly attempts to link specific computational properties of LLMs to specific capacities. 

On the other hand, Large Language Models might model a \emph{corpus}: that is,  LLMs model a collection of text, but not by possessing or modelling the capacities that gave rise to it.  Most deflationary takes on LLMs say something like this. The idea that LLMs are mere `stochastic parrots' \citep{BenderOn-the-dangers21} or imperfect compressions of the training corpus \citep{ChiangChatGPT23} fall into this category. So do more nuanced critiques tying LLM performance to the exemplars seen during training \citep{McCoyEmbers24}.  These responses explain the capabilities of LLMs primarily in terms of the richness of the training data, rather than specific human-like capacities.  

This paper will fall into the latter camp, though with (I hope) a relatively non-deflationary conclusion.  The argument will come in several steps of increasing speculativeness.  

First, I will argue (\textsection\ref{formatsec}) that cognitive science gives us reasonably good facts about the kind of computational  \emph{format} that supports cognition. Linguistic capabilities in humans rely on what I'll call \emph{supralinear} formats for computation: that is, formats for processing that outstrip mere linear ordering of components  Next, I'll argue  (\textsection\ref{invariantssec} and \textsection\ref{residualsec}) that transformers do not have the right format to be either using or modelling human capacities, because they support at best \emph{linear} formats for processing.   My target when talking about format will be the format of the  {residual stream} of the transformer, construed broadly enough to include the input and output embeddings. This is because I care about format insofar as it interacts with computation, and the residual stream (unlike the weights) changes throughout a forward pass---the residual stream is, if you like, where all the action happens. This argument will rely primarily on certain invariants of the computational architecture, but I will flesh it out with discussion of the details of processing in the residual stream.

Following that, I'll sketch a positive story (\textsection\ref{positivestory}) about what transformers are doing, focusing on the idea of shortcut automata suggested by \citep{LiuTransformers22}. If we take transformers seriously as doing sequence-to-sequence predictions (rather than just next token prediction), then their outputs don't fit particularly well with linguistic processing but dovetail nicely on a view where they have learned to emulate productive finite state automata. I then conclude (\textsection\ref{conclusion}) with why I don't think this is a terribly deflationary story. Language is not (just) a means for expressing inner state but also a kind of {technology} invented by humans that allows us to systematically transform bits of language given appropriate context. We have learned to use this technology in one way. LLMs have also learned to use it via very different means. 

A few caveats before I start. The focus on this paper will be on (very) simple LLMs: encoder-only transformers performing next-token autoregressive tasks (from now on, when I say `transformers' or `LLMs' I will refer to these specifically).  These are, arguably, the  simplest successful large language models, and it is that simplicity that makes them attractive for philosophical analysis. More sophisticated models obviously perform better, but the simplest models already look to be doing what more sophisticated models do. Examples come from the first-generation OpenAI GPT, and were generated directly in Python. 

Finally, while I will be concerned with the details of transformers, I am keen to avoid what \citep{MilliereA-Philosophical24-1} call the `re-description fallacy'. As they explain it \begin{quote} This fallacy arises when critics argue that a system cannot model a particular cognitive capacity, simply because its operations can be explained in less abstract and more deflationary terms. In the present context, the fallacy manifests in claims that LLMs could not possibly be good models of some cognitive capacity $\phi$ because their operations merely consist in a collection of statistical calculations, or linear algebra operations, or next-token predictions. \citep[9--10]{MilliereA-Philosophical24-1} \end{quote} I agree that this sort of move is too cheap. For one, it's self-undermining: no computer really does linear algebra anyway, because it's \emph{really} all just logic gates down there.  For another, \emph{tu quoque}: all you're doing is sloshing chemicals around.  One wants something more, and the something more is arguably more demanding the more deflationary you want to be. The argument in section \ref{invariantssec}  is meant to show one way to answer those sorts of challenges: focusing on mathematical invariants gives us a principled way to interpret the actions of the transformer.

\section{The format of thought}
\label{formatsec}

\subsection{Why format matters}

We represent the world. LLMs do as well. When we talk about representation, we can ask two important questions: what is represented and {how} it is represented. The latter question is sometimes phrased as asking for the \emph{format} of the representation.  Intuitively, format has something to do with the structure of the representation itself: the same content can be represented in different ways by structuring it differently. Different structures are better for different goals: a topological subway map is better for navigation, a textual description more suited to legislation, a cadastral map best for maintenance. 

In addition to familiar public representations, the notion of a format also plays an important role in theories of computation. Very basic algorithm design  cares a lot about the format of data, and much foundational computer science is devoted to subtly different formats considered at a very abstract level (see e.g. \citealp[Ch2]{KnuthThe-Art-of-Computer97}). To understand an algorithm we must also understand the format of the data on which the algorithm operates and vice-versa, a principle that  Marr calls ``the duality between representations and processes'' \citep[331]{MarrVision:82}. 

While abstract, some high-level differences in computational format are familiar even to casual users: \texttt{.svg} and \texttt{.mp} are two different formats for representing pictures. Each can represent the same picture, but each with different pros and cons: \texttt{.svg}, with its vector-oriented operations, scales cleanly. 

Given the link between cognitive science and the computational theory of mind \citep{NeisserCognitive67, PylyshynComputation84}, it shouldn't be surprising that cognitive science has cared about format too. Most cognitive theories also imply a certain format for cognitive processes, and the format of cognitive processes has  empirical significance. To take a classic example, empirical results on the mental rotation of three-dimensional objects \citep{ShepardMental71} have suggested that the format of visual imagery is something like a picture or an icon. Ensuing debates focus on whether the data demand such an interpretation of the format, or whether some other formal might support the same results  \citep{PylyshynReturn03}.  Similarly, the classic debate over the \emph{language of thought} hypothesis \citep{JFLOT,Quilty-DunnThe-best23} is, at heart, a question about the format of the representations used in human cognition and how that format shapes the way that information enters into computation.

\subsection{Content-neutral format}
The idea of a format is used in a variety of ways \citep{MolloThe-formats23}. Some discussions of format presuppose that formats also have a certain kind of representational content. In ordinary uses of the term  `map', for example, something is a map because it has a certain kind of format \emph{and} it represents spatially arranged locations. A map may misrepresent, of course, or it may represent a fictional place. But such claims these claims themselves assume that a map is what it is in virtue of having a certain kind of representational content. One way to put this is that the ordinary sense of map comes  equipped with a semantics that lets users of the map link it to the world. This additional semantics can make it hard to distinguish properties of the format from additional interpretive structure that \emph{users} of a format bring to it.  

On the other hand, some examples of format---and particularly those that appear in computer science---use a \emph{content-neutral} sense of format. A \texttt{list},\footnote{I use typewriter font when discussing data types to make clear that I'm talking about the type rather than the intuitive meaning of the term. In some places I'll also use the font for operators on that data type; it will be clear enough which is which. While the data types themselves are meant to be idealisations, in practice I'm mostly using examples  from Python.} in this bare sense, is just an ordered sequence of some items or other. We can use a list to organise information about subways, words, or groceries. Importantly, we can use and reason about a \texttt{list} despite it representing nothing at all.  We can certainly reason about \texttt{list}s  without caring whether or what they represent.  

Another way to put the point is that bare formats are completely general, in the sene that they come with no semantics that links instances to the world. However, bare formats still have an \emph{internal} semantics, which defines a format by showing what stays the same across different operations.\footnote{See \citep[33ff]{Cantwell-SmithOn-the-origin96}.  \citet{KleinWhat21} refer to this as the difference between exoteric and esoteric semantics.} 
So, for example, an \texttt{int} is a single entity that can take one of a range of values. A \texttt{list} contains a finite number elements of and a well-defined complete ordering over them. That means you can do things to a \texttt{list} like traverse it in order. If its elements can be compared, you can sort it. A \texttt{set}, by contrast, also has elements, but without order; it does not make sense to sort a \texttt{set}. There are other, more subtle differences. Adding an element to a \texttt{list} always happens at a certain place; not so with a \texttt{set}. If you add an element to a \texttt{list} you are guaranteed to get a new, distinct list. That is  true of a  \texttt{set} only if the element is novel. 

Thinking about operations and invariants is useful is because it allows us to talk about content-neutral formats without without having to worry about what operations are operating \emph{on}: that is about the vehicles for representation.  For example, a classic textbook  on programming \citep{AbelsonStructure96}  discusses the idea of a \texttt{pair} thus:\begin{quote} We never actually said what a pair was, only that the language supplied procedures \texttt{cons}, \texttt{car}, and \texttt{cdr} for operating on pairs. But the only thing we need to know about these three operations is that if we glue two objects together using cons we can retrieve the objects using \texttt{car} and \texttt{cdr}. That is, the operations satisfy the condition that, for any objects x and y, if z is \texttt{(cons x y)} then \texttt{(car z)} is x and \texttt{(cdr z)} is y. Indeed, we mentioned that these three procedures are included as primitives in our language. However, any triple of procedures that satisfies the above condition can be used as the basis for implementing pairs. \citep[\textsection 2.1.3]{AbelsonStructure96}\end{quote}  In other words, once we've set up what you can do to a \texttt{pair} and specified the appropriate invariants, we've specified the data type.  Indeed, the passage ends with the  claim that ``This point is illustrated strikingly by the fact that we could implement \texttt{cons}, \texttt{car}, and \texttt{cdr} without using any data structures at all but only using procedures.''\citep[\textsection 2.1.3]{AbelsonStructure96}. 

As a general principle, I suggest, we distinguish bare formats by (a) the core things you can do to instances of them, and (b) what you can expect to stay the same when you do those things. As a shorthand, I'll say that content-neutral formats are defined partly by the \emph{invariants} that they support.\footnote{ I won't make much of the theory of invariants, but I make more of it in my \citep{KleinComputational26}, which is  in turn inspired by Marr's under-appreciated group-theoretic discussion of computational individuation \citep[22]{MarrVision:82}.  The overall approach to computation is ultimately meant to be an algebraic one \citep{EilenbergAutomata74a}; For a proper formal treatment of types in particular, one might look to \citet{GuttagAbstract77}, and in particular his idea that data structures can be individuated by the invariants of the operations defined upon them. } 

Conversely, we can also get some insight into the sorts of formats that are used in a computation by looking to see if the right sorts of invariants are supported. Invariants are properties that primarily appear at  \citet{MarrVision:82}'s computational level. Format appears at the algorithmic level, and the  computational constraints the algorithmic.  Hence by looking at the invariants of an architecture, we can sometimes determine details of format. 

\subsection{Derived formats}

I  have mostly focussed on simple formats. However, we can use simple formats to create more complex data types out of the basic ones \citep{WirthAlgorithms76}. So, for example, a \texttt{Graph} can be built\footnote{This is, to a first approximation, how it's done in the Python package \texttt{networkx} \citep{HagbergExploring08}.} by starting with a basic key-value \texttt{dict}. Take, say, a \texttt{dict} of \texttt{dict}s -- one of nodes, one of edges. Each of these may in turn contain \texttt{dict}s of attributes. To make it do what a graph is intuitively meant to do, one needs to add a bunch of procedures that keep in place the right guarantees: adding an edge also automatically adds the appropriate nodes (for example). There may also be other convenience functions to keep track of things explicitly represented (like the neighbours of a node).  Call this a \emph{derived} format. 

Derived formats are important for understanding computations. When we talk about (e.g.) brains containing maps, the map-like format should probably be understood as derived out of more basic operations provided by cortical neurons.  From the point of view of \emph{theory} of formats, though, derived formats work just the same as basic ones. Indeed, the  process of creating derived formats is just an extension of the operations-and-invariants logic. 
One builds up a new formats by combining together or restricting (subtyping) more basic formats; the point of restriction is to add both new operations and new guarantees to the format. 

Derived formats do add one conceptual wrinkle, however, that's worth making clear. Recall that the entire discussion so far has been about content-neutral formats. Derived formats are still content-neutral: a \texttt{Graph} is a bare structure, and needn't represent anything. However, in many cases it can become difficult to distinguish between \emph{representing} structure and \emph{representational} structure. Merely taking a picture of a graph, or something like a graph-like structure, is not enough to actually create a \texttt{Graph}. Why? Pictures do not have the right sorts of operations, and do not  uphold the right invariants. So while a picture can carry the same information that a \texttt{Graph} embodies, we want something more out of formats (Compare here \citet{HaugelandRepresentational98}'s reflections on representation versus mere recording.)  

Derived formats will play a minor role when we return to transformers. To foreshadow the argument a bit: there is a sense in which transformers \emph{must} represent facts about language, linguistic structure, and even linguistic processing. That is guaranteed by the surface data. Our question is whether this is merely a fact about the content of the representations that they use, or whether their representations have the same format (basic or derived) as the representations we use.

\subsection{Representational families}

Discussions of cognitive format occur at different levels of abstraction. Many debates in cognitive science are about the details: two theories can agree that spatial orientation in humans requires map-like representations, but differ on how those maps work and what they must represent. \emph{Philosophy} of cognitive science is often more concerned with what \citet{HaugelandRepresentational98} calls `representational genera'---very broad categories like languages, icons, maps, and so on. 

We can also take one step further up---to the representational \emph{family}, as it were. At the level of the family we may draw a rough three-way distinction between \emph{sublinear}, \emph{linear}, and \emph{supralinear} formats.  Sublinear formats impose no particular ordering on the items in the format---that is, no operations are order-sensitive, and nothing about order is guaranteed. Sublinear formats include basic elements like \texttt{int}s and \texttt{float}s and  formats like \texttt{set}s and \texttt{dict}s. A linear format like a \texttt{list} or a \texttt{tuple}, by contrast, allows order-sensitive operations and guarantees a strict ordering between subparts. Finally, a supralinear format (like a \texttt{graph} or an \texttt{array} or a structured object in an object-oriented language) has `more' structure than just a linear one (perhaps because there is more than one way to order elements, perhaps because elements admit of a partial ordering, perhaps because different operations apply to different elements). 

The distinction between representational families is relatively easy to make in the case of bare formats because representational families will have different invariants. The fact that a \texttt{list} admits of a complete ordering over the elements is baked into what it is to be a list: when you add something to a list you add it \emph{at a position}, and nothing else can be at that position, and the position it is at will change systematically if you add other things, and you're guaranteed to get that position back when you go looking. 

The level of representational family is also a useful one for cognitive science, because many classic debates about the format of cognition really come down to debates about the representational family needed to achieve certain competences. The conclusion is nearly always that something supralinear is required.  Most notably, Chomsky's argument against Markov models of language \citep{ChomskyThree56} can be read as concluding that no merely linear format can be sufficient for syntactic processing.  Many of the classic debates above were  about representational genera \emph{within} the supralinear family---that is, whether humans use a language of thought or an icon to rotate pictures---which implies that only supralinear formats will do. Conversely, these debates suggest that  in many cases we can find sufficient constraints at the computational level to at least narrow down candidate formats to a specific representational family. 

This also provides us with a possible handle on questions about transformers and their capacities. From here out I will assume that human cognitive performance requires computations that use supralinear formats. The primary question will therefore be whether there is evidence that transformers can support such formats. The source of that evidence will be invariants transformers, and particularly those that depend on the transformations of the residual stream.

\section{Two invariants of the residual stream}
\label{invariantssec}

\subsection{The residual stream}

I will be concerned with the format of the so-called \emph{residual stream}, which caries data from step to step in a transformer. I start with a very high-level overview of the transformer architecture, focusing on features relevant to the argument There are better introductions to be found.\footnote{See e.g. \citep{VaswaniAttention17,AlammarThe-Illustrated18,MilliereA-Philosophical24-2}}. This is meant to emphasise features that will be key for the argument.  

\begin{wrapfigure}{r}{0.2\textwidth}
\begin{center}
    \includegraphics[width=0.2\textwidth]{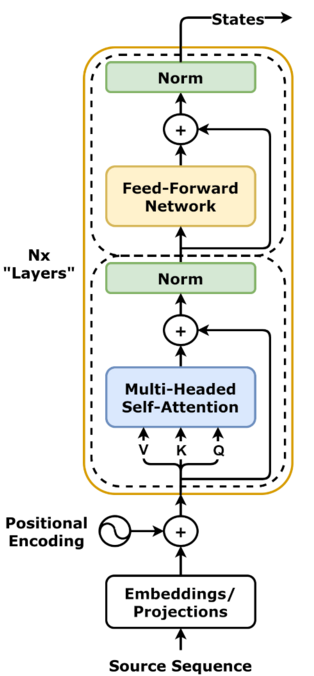}
  \end{center}
  \caption{A depiction of a simple encoder-only transformer. Image courtesy wikimedia commons (CC BY 4.0).}
  \label{transformerpic}
\end{wrapfigure}

Figure \ref{transformerpic} shows a simple encoder-only transformer. Data flows from bottom to top. The central motif (a multi-head self-attention block followed by a multilayer perceptron block) repeats a number of times, each taking the output of the last block as input. After a number of these layers, the output can then fed into a linear decoder (not pictured) and a softmax function, which gives a probability distribution over possible tokens.  For text generation, the final token is chosen using the resulting distribution, appended to the original input; the process is iterated until text of the desired length is reached.

The residual stream is  represented by the black arrows that reconnect after each of the blocks in figure \ref{transformerpic}.  Insofar as the transformer is representing anything specific to the task, that happens in the residual stream. Everything else---the weights in the Attention and MLP blocks, the various embeddings and encodings---are all fixed by training. If they represent, they represent something general learned from the corpus.

There are two different types of basic transformer architectures that differ in a single operation performed on the residual stream. Some transformers let  any position in the residual stream attend to any other position. This is common when doing masked token prediction anywhere in a sentence (as happens with BERT \citep{DevlinBert:18}).  Other transformers---including the canonical one in \citet{VaswaniAttention17} and most autoregressive LLMs---only allow tokens to attend to earlier ones in the sequence. Formally speaking, this is accomplished by a single additional operation, masking, which makes the attention matrix lower triangular before multiplying by the value matrix. I will refer to the first sort of transformer as a \emph{unmasked} transformer and the latter as a \emph{masked} transformer. 

The format of the residual stream will be determined by the specific operations of the transformer blocks and the guarantees that these support. Much work on mechanistic interpretability focuses on a relatively fine-grained look at these operations. I have suggested above that the mathematical invariants of transformers might also carry important information about the format used.  The two types of transformers have two distinct (but ultimately related) invariants. I will consider each in turn. 

\subsection{Permutation Invariance}
\label{perminvariance}

Unmasked transformers are what is known as \emph{permutation invariant}.\footnote{A brief technical note: I am eliding the fact that completions themselves are stochastic. Everything I say is really true of the probability distribution over final tokens; for exposition, I have assumed a fixed seed, and so similar completions. It is actually the probability distribution that matters for the argument.}  Given a transformer $T$, an input matrix $X$, and a permutation matrix $P_\pi$, then $T(P_\pi X) = P_\pi T(X)$.  (See appendix \ref{PEProof} for a proof.)  In other words, if we feed a transformer some input and then permute the tokens of the {output}, what we get is the same as if we permuted the tokens of the input in the same way.

This is a remarkable property when first encountered, and worth  more attention than it is often given.  It appears to imply the following: Suppose we start with ``cats eat fish and'' and $T(X)$ gives the plausible completion ``rats''.  Let  $P_\pi$ swap the first and third rows of a matrix. Permutation invariance implies that if we give $T$ ``fish eat cats and'', it should also return ``rats''. Indeed, even a completely ungrammatical permutation will work: if we feed the transformer ``and fish cats eat'' it will return ``rats''. 

Of course, LLMs do not actually do this. It's a good thing they don't. \emph{Language} is not permutation-invariant: the order of constituents matters! Given the facility of LLMs with language, something else is going on. 

The typical answer is that the input to an LLM is actually the embedded input $E+P$, where $E$ is the token embedding and  $P$ is a positional encoding matrix (see appendix \ref{positionalencoding} for more detail and reflections). Positional encoding adds additional information to the individual constituents of what would otherwise be an unordered multiset. As a crude example, if we represent our sample sentence above as: 
$$\{\{\text{`cats'},0\},\{\text{`eat'},1\},\{\text{`mice'},2\},\{\text{`and'},3\}\}$$
then permute the elements of a representation of this set, we preserve information about ordering. That is, the same sentence can be represented equally well as: $$\{\{\text{`mice'},2\},\{\text{`eat'},1\},\{\text{`cats'},0\},\{\text{`and'},3\}\}$$ because it is,  of course, just the same set.

The addition of the position encoding, then, is often treated as a boring implementation detail. In  \citeauthor{MilliereA-Philosophical24-2}'s insightful two-part treatment of the state of the art in LLM interpretation, for example, positional encoding is mentioned exactly once, in a figure caption \citep[p11 fig 2]{MilliereA-Philosophical24-1}.  But the fact that position information \emph{is} needed shows us something very important about the core operations of the basic transformer architecture. 

 Indeed, it is worth emphasising that  positional encoding  does not change anything about permutation invariance itself. The basic architecture is always permutation invariant---position encoding just gives the transformer access to representations of position information to which it would not otherwise have access.

\subsection{Substring invariance} 

Masked transformers have a distinct property I will call \emph{substring invariance}. Informally speaking, substring invariance says that if we give a transformer a long strong, what it does on and to the initial portion of that string is the same as what it would do if we just fed it that initial portion on its own. So if I give a transformer ``When lions eat humans they get a bad'', then the first four tokens of the output, and the first four rows of each position in the residual stream,  will be the same that I would get if I fed it ``When lions eat humans.'' 

Figure \ref{AllLogits} illustrates this property. It shows the  shows the transformation of a sample sentence by an early transformer model, decoding the top logit for each position. Of course, when we iterate it is only the final token that's appended to the original sentence (not the transformed sequence).  However, note that the token in row 12 (the top logit of the output) is a reasonable completion for the input that consists of just the columns up to that point.  Indeed, the first $n$ columns match what one would see if one input only the first $n$ tokens.

\begin{figure}[h]
\begin{center}
\begin{tabular}{l l l l l l l l l }
0  & when      & lions     & eat       & humans    & they      & get       & a         & bad       \\
1  & exa       & lions     & eat       & humans    & they      & get       & a         & bad       \\
2  & arias     & lions     & eat       & humans    & they      & get       & a         & bad       \\
3  & martel    & lions     & eat       & humans    & 'd        & get       & lot       & bad       \\
4  & martel    & lions     & eat       & humans    & 'd        & get       & lot       & bad       \\
5  & martel    & lions     & eat       & humans    & 'd        & get       & lot       & bad       \\
6  & morrow    & lions     & eat       & humans    & 'd        & get       & lot       & bad       \\
7  & morrow    & lions     & eat       & indiscri  & tess      & get       & lot       & bad       \\
8  & ouses     & lions     & dine      & indiscri  & emit      & rar       & lot       & taste     \\
9  & trait     & lions     & torto     & ,         & 're       & pees      & lot       & taste     \\
10 & ouses     & pees      & excrement & ,         & 're       & vacc      & lot       & taste     \\
11 & penc      & are       & torto     & ?         & can       & vacc      & taste     & taste     \\
12 & he        & and       & their     & .         & do        & a         & little    & feeling   \\
\end{tabular}
\end{center}
\caption{OpenAIGPT on a sample sentence. Each row represents the decoded residual stream after the operation of full block. The continuation is given by the final column in the final row. Highest-probability token for each row shown. }
\label{AllLogits}
\end{figure}

Substring invariance is a a function of the masking operation: intuitively, the mask makes sure that each position is only affected by earlier tokens, giving zero weight to later ones. So any tokens that are added after a substring cannot have an effect on the operations on earlier tokens. (See appendix \ref{substringproof} for a formal proof and demonstration; note that in practice many masked transformers are only \emph{approximately} substring invariant, but the departure is small enough that it won't make a difference for the argument.) 

Formally speaking Let $X[n]$ refer to the first $n$ rows of a matrix $X$. A transformer $T$ is {substring invariant} just in case for any $n$, $T(X[n]) \approx  T(X)[n]$.  Indeed, we can actually say something stronger: let the residual stream after the $m^{th}$ step on input $X$ be designated by $R_m(X)$. Then for any $m$, $R_m(X[n]) \approx R_m(X)[n]$. So not only is the output substring of a masked transformer invariant, but the output after each block of a masked transformer block is substring invariant as well. 

Substring invariance should not be surprising. As \citet{VaswaniAttention17} put it,  ``We also modify the self-attention sub-layer in the decoder stack to prevent positions from attending to subsequent positions. This masking\ldots ensures that the predictions for position $i$ can depend only on the known outputs at positions less than $i$.'' (p3) Hence substring invariance is just a consequence of the masking operation working as expected.

\subsection{The relationship between the two}

Substring invariance and permutation invariance are, in a straightforward way, mutually incompatible. Nevertheless, there is an important relationship between the two.  

Masking is just one additional operation added to unmasked transformers. This perspective emphasises the fact that the other core operations of a transformer---everything other than masking---are agnostic about the position of elements in the input. Indeed, in simple unmasked implementations, masking just throws away information that the transformer had already computed. So the action of masking must be understood against a background of an architecture that is otherwise indifferent to the rows in which information is represented. Further, masking is sensitive \emph{only} to actual position, not to the content of those positions (including any representation of position information). 

A compact way to put this point is to note that masked transformers are still permutation invariant \emph{with respect to earlier tokens}. What comes before a token matters, but not the order in which it is stored. This is why even masked transformers need a position encoding: they can work with initial position information only insofar as it is represented in the content fed to the input.

\section{The format of the residual stream and the format of thought}
\label{residualsec}

We are now in a position to draw a number of conclusions about the  representational genera  of the residual stream,  of both basic and masked transformers. This section takes a circuitous path;  here are the highlights up front. First, I'll argue (\textsection\ref{basictransformersformat}) that the basic format in unmasked transformers appears to be extremely lightweight; if it has any further structure, it must be a derived format built on top. Then I'll argue (\textsection\ref{maskedtransformersformat})) that masked transformers have linear formats, but are restricted to having \emph{only} linear formats. I'll return briefly then to unmasked transformers and suggest that the same applies to them (that we should view the difference as essentially that between singly and doubly linked lists); he conclusion will be that transformers probably cannot support any format with supralinear structure.  The interim conclusion (\textsection\ref{interimconclusion}) is that transformers are not processing language in the way that we process language. 

\subsection{Unmasked transformers}
\label{basictransformersformat}

Begin first with unmasked transformers. Take the residual stream at any step. Start with what we might call the basic format of the residual stream (call it $R$). Just looking at the invariants we have established, what can we say about $R$?

When we zoom in to look at the basic operations actually performed, we can say three things.  First, $R$ is a collection of vectors.\footnote{When I say $R$ is built of vectors, I use  `vector' to refer to the computational format rather than the mathematical object that is readily represented by the object: roughly, a vector is a fixed-length list of floating point numbers. In particular programming languages there would be specific classes, the names of which (arrays, tensors, etc)  aren't necessarily less confusing, so I'll stick with `vector'.} Second, $R$ imposes no intrinsic relationships between these vectors.

Third, the basic operations performed on instances of $R$ don't seem capable of adding further structure on their own, because they can't break permutation invariance.  There are four types of operations performed on the residual stream in unmasked transformers (See appendix \ref{fourtypes} for more detail). Two of them treat each individual vector in $R$ independently. The other two are all-to-all relationships. So either the operations treat each vector independently, or lump together the whole set. There is no special significance to which vector is where, and no privileged relationship between any particular pair of vectors. 

Conversely, this means that permutation invariance should not be particularly surprising.\footnote{What is perhaps more surprising is that permutation invariance holds for the training phase as well as for the use of transformers.  \citet{XuPermutation24} show that with a few additional constraints,  permuted input can be trained just as well as ordinary input. I won't focus on that property here though.}  Permutation invariance shows that none of the four basic operations can introduce privileged relationships between any two rows of the residual stream.  When we look to the invariants that would  constrain $R$, permutation invariance puts a hard constraint. We can't mark off some rows as intrinsically special, either in their own right or in relationship to other rows. 

Formally, $R$ appears to be a kind of of \texttt{mulitset}---it is a container for a fixed number of objects, with no order, that behaves like a set except that there can be multiple copies of the same elements. Less formally, the residual stream is effectively a \emph{sack}.  It has the format you'd use to represent a bag of scrabble tiles (if scrabble tiles were 768-dimensional vectors). The operations performed on it are ones applied to each individual tile, sometimes comparing tiles to one another, more often to target sacks of tiles, and using the results to edit each tile a little bit. From this point of view, the format of $R$ is extremely unstructured: indeed, it is \emph{sub}linear, without any intrinsic ordering. 

To be fair, I have focused only on the most basic invariants that constrain $R$.  It is possible to use unstructured formats to build more structured ones, so long as we have the right kinds of operations performed consistently at each step. And there is an obvious  mechanisms by which this might happen: the original positional encoding. Positional encoding adds order information to the original token information, and makes it available for manipulation. So we have, at least in principle, a way in which position information might be added, independently manipulated, and used to create more structure during the course of an unmasked transformer's operation. 

That said, it seems that whatever is going on is more complex than simply using positions over the course of a transformation. The original position information itself isn't obviously preserved in the course of processing (see figure \ref{posfig} in supplemental appendix \ref{positionalencoding}). That's not necessarily problematic: we might view each block of the transformer as having a residual stream with a new format, for example, with each step partly serving to bring the format from the original linear format to something else more useful.\footnote{Thanks to EJ Green for raising this issue, and also for making things a lot harder for me.}  This would be akin to what happens when (say) a Python package like \texttt{NetworkX} reads in a list of nodes and transforms it to a proper graph. 

So whether this is enough of a constraint to give rise to an actual format depends a lot on the details, and in particular whether the actions of the blocks in a unmasked transformer consistently preserve some kind of format-related information. That is an open question, and a difficult one to answer.  Rather than come at it directly, let's turn to masked transformers.

\subsection{Masked transformers}
\label{maskedtransformersformat}

Unmasked transformers do not distinguish between the position of tokens. Masked transformers do: there is a clear distinction between earlier  and later positions in the residual stream at the attention step. Indeed, it is trivial to construct a strict ordering over positions in terms of which positions can influence which others.  The basic operations still do not care about position, but masking places an additional constraint, and thus a new invariant, on the format $R^\prime$ of the residual stream of masked transformers: we can reliably distinguish between earlier and later tokens of $R^\prime$, and so expect differential effects given different positions. 

On the one hand, this means that the format $R^\prime$ is more structured than the $R$ of unmasked transformers. $R^\prime$  is at least linear. We can talk intelligibly of earlier and later positions in instances of $R^\prime$, because for any pair of instances of tokens, masking ensures that we can tell that one came before the other. 

On the other hand, I suggest that $R^\prime$ is also at \emph{most} linear. That is, the additional constraints placed by masking also make it hard to see how any deeper structure might be found without violating substring invariance.  For substring invariance puts strong constraints on the kind of structures that we might think to construct. Take some sequence of input tokens $X = t_1,t_2,...t_n$. Substring invariance says that whatever will be done on $t_1,t_2,...t_{n-1}$ on its own, \emph{including any invariants that would give rise to a derived format}, will also be done to the first $n-1$ tokens of $X$. Later tokens do not affect earlier tokens. 

This would seem to preclude, for example, the construction of structures like a \texttt{tree} or a \texttt{graph}. Even if the initial input to a tree or a graph is linear, part of the key feature is that one can make substructures that bind later bits of information to earlier ones.  Syntactic parsing ought to work like this. Consider the two sentences ``To fish, you need a rod'' and ``To fish, you are a giant.'' They have the same two initial words, but in the first sentence `fish' is a verb and in the second it is a noun. That is disambiguated only by later information in the sentence, with corresponding differences in building up a syntactic tree.   Yet substring invariance precludes any such operations: later tokens cannot be bound to earlier structures. 

Now, this is not  an entirely knockdown argument. With sufficient cleverness, one can (for example) construct a \texttt{tree} from a \texttt{list} in a way that preserves substring independence. The trick, however, is that later positions in the list must contain information about the structure of the whole tree up to that point---or, more generally, we must allow for any position in the residual stream to carry complete information about the format, from the point of view of that position, \emph{on its own}. Thus the longer the input, the more demanding the processing done on later positions. 

I see no obvious mechanisms by which transformers could do this, however. Here we may return to a subtle interplay between permutation and substring invariance. Permutation invariance reminds us that the \emph{other} basic operations, including the operations of the attention and the MLP blocks,  are performed in a position-independent manner. Again, each operation is applied to each token; the only thing masking does is throw out the result of some of those operations before further processing. This would mean that to get further derived structure, each position in the residual stream, and each operation, would have to be capable of supporting  arbitrarily complex operations and structures. That is a tall order.  

Indeed, having established this, I think it provides strong abductive evidence that even basic transformers are limited to at most linear format.  I noted that the core format of the residual stream appears to be sublinear, but that this does not preclude a supralinear format being built on top. I now want to suggest that, in fact, we should think of the format of the residual stream even in basic transformers as no better than linear. The argument is mainly abductive. Other than the masking step, there is no obvious reason to think that basic and masked transformers work in a substantially different way.  Absent that, we ought to think that what masked transformers do with mere linearity, basic transformers do as well. Conversely, the addition of position encoding gives a natural way in which the residual stream of unmasked transformers might implement a linear structure with a sublinear basic format. 

\subsection{Interim conclusion}
\label{interimconclusion}

Let's pause briefly to take stock. I began with the question of whether transformers modelled human linguistic capacity or something about the corpus. One way to get at that question is to look at the format of the representations used by transformers. Cognitive science says that humans use supralinear formats. Format implies invariants. The invariants of transformers suggest at most a linear format for the residual stream, which is where the computation happens. So we have good evidence that transformers do not process language in the way we do. The first path is blocked. The remaining sections will discuss the second option: that transformers model the corpus. 

One could, of course, get off the boat at an earlier step, and read the preceding as a kind of \emph{modus tollens}: if transformers can do so much with so little, maybe so can we! 

I am not going to pursue this line. For one, I think the cognitive science evidence is substantial, covers numerous areas, and is not so easily dismissed. For another, human brains are substantially constrained by powerful energetic constraints \citep{SterlingPrinciples15}. Formats allow computational processes to do more with less, and evolution always demands we get by with less. 

And finally, I think it is to overlook important facts about why transformers are so powerful. For I suspect that transformer architectures are flexible and powerful precisely because they \emph{don't} bother to use structured representations.  The state of the art before transformers can be seen as various ways of trying to introduce structured representations into the architecture  in ways that let it capture the structure of the target domain. Convolutional deep neural nets for image processing restrict all-to-all connectivity in order to allow the network to exploit spatial contiguity. Recurrent networks for language modelling present elements sequentially and use the effect of earlier computations to alter processing of later ones, trying to build a format with sensitivity to syntactic structure. 

The cost, of course, is that the resulting architecture is tuned to particular domains.  Transformers, by contrast, are very general because they eschew format in favour of representational content: the structure of the domain gets captured by the operations of the transformers themselves. So, for example, \citet{CordonnierOn-the-relationship19} argue that the early attention layers of image transformers do something very much like the initial convolution operations in convolutional deep neural nets. Rather than build in those convolutions as part of the architecture, though, they're just one of many learned transformations. 

So I think we should think that transformers really are doing something different with language than what we're doing.  Now, this might sound too deflationary, and one might be skeptical of it as a conclusion----not so much on specific grounds as because it seems impossible for LLMs to do what they do with a merely linear format.  With that in mind, I now turn to an---admittedly speculative and big-picture---story about how linearity might be enough to do the job of modelling a corpus.  It is to this question I now turn.

\section{Linearity and Automata}
\label{positivestory}

\subsection{Automata and state-strings}
I have focused so far on the production of the output token. However, it's worth noting that the official task of a transformer (even a feedforward, encoder-only transformer) is a sequence-to-sequence transformation: that is, a transformer always takes a full input sequence and returns a full output sequence. Since only the final token of the output sequence  is appended to the input,  the remainder of the output is often ignored. Yet it is often striking how little the full output sequence resembles the input sequence. 

Return to figure \ref{AllLogits}, which was used to illustrate subsequence invariance. What figure \ref{AllLogits} makes clear is that the  transformer is \emph{not} preserving information about the original sentence (or some subpart) and simply trying to guess the last token. Instead, the sequence that is produced from the original sequence is \emph{the sequence of final guesses for each substring}.

 That is a very odd thing to interpret in linguistic terms.   However, there is another way to interpret what's going on. Briefly: we could view transformers as modelling automata that can produce the input corpus given appropriate input.

 Start with the idea of an \emph{automaton}. Automata are models of computation that  proceed in discrete steps, at each step  receiving an input from a finite alphabet of symbols,  transitioning to one of a finite set of states,  possibly altering the state of a (possibly unbounded) memory, and possibly producing some output along the way. The simplest sort of automata, so-called Finite State Automata (FSAs) take an input at each step, transition state deterministically, and have no further memory. I'll focus on FSAs for now, though I'll come back to this restriction below. 

For any automaton $A$, we can define a function $f_A$ that takes an input string $I$ and returns a sequence $O$ of states that $A$ goes through as it consumes $I$. Note that this will be different to the function that $A$ itself computes or the output it produces: not all automata produce output, and in many cases you can't work backwards from output to sequence of states. 

Now, $f_A$ has two useful properties. First, the sequences produced by $f_A$ must be subsequence invariant: if we add a new symbol to our input, $A$ will do all the same things up to that input before something new happens. This is another reason to distinguish $f_A$ from the output or function computed by $A$: depending on how you define the output $A$, the addition of a new symbol might make an arbitrarily complex change to the output. (Consider the task of detecting a well-formed string, where the addition of a single token can completely change the answer.)   However, the addition of a new symbol will always make an incremental  change to the output of $f_A$.  

Second, it is often possible to calculate $f$ using fewer steps than the $A$ itself  would take, especially if we can perform parallel operations on the input string. As a trivial but instructive case, consider the simple reset automaton $\bar{\mathbf{2}}$: it has two states $A$ and $B$,  transitioning to $A$ when it reads a `0' and to $B$ when it reads a `1' (regardless of current state). Given an input of $n$ symbols,  $\bar{\mathbf{2}}$  will run for $n$ steps. However, $f_{\bar{\mathbf{2}}}$ can be calculated in a single parallel step,  regardless of the input size: replace every `0' in the string with an $A$ and `1' with a $B$. 

Following \citet{LiuTransformers22}, I'll call this a \emph{shortcut} solution. A shortcut solution is a calculation of $f_A$ which takes advantage of parallel processing to use substantially fewer steps than $A$ itself would on its inputs. 

Return now to LLMs. Suppose we take LLMs to be primarily concerned not with language as such, but rather with \emph{calculating the input-to-state function $f_A$ of some corpus-producing automaton $A$}. This requires, on the face of it, only facility with linearly structured inputs and outputs. It would also neatly explain subsequence invariance. It also leaves otherwise entirely open what the underlying automaton would have to look like. Importantly, it need not look anything like the automaton that underlies \emph{human} competence with language: so long as we restrict ourselves to finite strings, there are indefinitely many automata that can consume strings of input tokens in the appropriate way. 

\subsection{Transformers and emulation}

Of course, that doesn't yet say anything about how transformers manage to do what they do. \citet{LiuTransformers22} have demonstrated that in a great number of cases  there ought to be shortcut solutions to calculating $f_A$ for any given $A$. Furthermore,  the transformer architecture is well-suited to learning shortcut solutions.   I present the idea generally and informally; see appendix \ref{coveringappendix} for a number of details.

The key to the claim is an unexpected application of the Krohn-Rhodes theorem \citep{KrohnAlgebraic65}, which states that any automaton can be emulated by a cascade of very simple automata. Representing the transitions of such cascades can often be done very rapidly in parallel----as \citet{LiuTransformers22} note, in many cases, by a \emph{single} transformer block. 

Appendix \ref{coveringappendix} has more details of the argument. I do not think we need to lean on the details here, however, (though they are important).  For, I suggest,  work like \citet{LiuTransformers22} suggests a clear alternative picture to how a substring-invariant parallel feedforward architecture like a transformer can handle autoregressive-style tasks. First, we take some $A$ that can produce exemplars from the corpus given initial statements as input (and, given the nature of the training, $A$ should produce other corpus-like sentences given an unseen initial fragment).  A transformer is fundamentally concerned with calculating $f_A$. Each input corresponds to a change in the state of $A$. The existence of shortcuts implies that one can sometimes figure out the corresponding states with minimal information about preceding ones. In other cases, the information about the earlier states in the calculation can help whittle down the possible number of states at a location.\footnote{This may be why, as \citep{nostalgebraistinterpreting20} points out, that a lot of the work of settling on a possible distribution for the output token within the first few layers. After that, the guess of token appears to be progressively refined, and that refinement requires increasingly inscrutable use of the other slots in the residual stream. Note that the explanations in terms of (e.g.) self-attention are still valid in each of the subsequent layers. (That is also why it makes less and less sense to interpret self-attention in those layers as self-attention to \emph{other parts of a sentence}, as opposed to just an all-to-all comparison in the earlier sack of tokens in their own format.) }

 To summarise, there are really two ways to capture the structure of linguistic production: one that is complex and involves highly structured formats, and another that focuses only on the linear production of new words in light of the preceding ones. The former requires supralinear formats. The latter can be captured by a model that only needs to represent linearly, because the serial transitions of the underlying automaton is a linear process. Transformers perform the latter task, which is why they are best understood as modelling the corpus, not our underlying linguistic capacities. 

\subsection{Some caveats} 

The argument above is speculative, and one might worry that it makes some rather large assumptions. For starters, the focus on FSAs might seem like an odd choice. The celebrated early results of  \citet{ChomskyThree56,ChomskyOn-certain59} suggest that computations can be put on a hierarchy, that FSAs are relatively low in that hierarchy, and that at least some human cognitive abilities require more powerful architectures. This is  important when one (say) goes to compare natural intelligence across phyla \citep{KleinComparing24}.  

Nevertheless, I think there are a number of things one can say. Note first that, the argument only needs some set of finite strings from the corpus to be \emph{modelled} by an FSA. Furthermore, as many (including  \citealp{LiuTransformers22}) have noted, the limited context window for transformers means that we only need to model linguistic competence for finite strings---the classic Chomsky arguments all concerned the possibility of unbounded strings, which is a much harder problem. (Conversely, one lesson from transformers might be that, given many of our utterances are in fact quite short and context-dependent, emulating what we do is in fact much easier than one might have supposed.)  

Now, the problem of emulating linguistic competence on finite strings can be trivially solved by the equivalent of a lookup table---for $n$ slots and $t$ tokens an FSA $n^t$ states would be sufficient. That is, of course, absurdly large; the surprising lesson of transformer architectures might be (a) that a much smaller FSA would do, and (b) that the problem of calculating $f_A$ for that FSA admits of interesting parallel  solutions. 

A related point is that FSAs are memoryless: they have no storage either for local computations or for the results of previous computations. But that is, of course, what makes transformers such a good match for emulating FSAs: they are \emph{also} memoryless. Indeed, the fact that transformers do not learn anything new after the training stage simplifies the problem somewhat. The lack of learning or of meaningful memory further restricts the class of sentences that we have to model to those which involve finite strings and no influence of previous utterances. On the other hand, this suggests certain limitations to transformers that stem from the traditional limitations on FSAs. Anecdotally, for example, many LLMs struggle with string reversal tasks (such as producing or detecting palindromes). String reversal by an FSA is much less efficient than by automata with more structured memory. Similarly \citet{LiuTransformers22} note that there ought to be certain input transformations that do not admit of shortcut solutions\footnote{In particular, when the transformation semigroup of the underlying automaton has a non-solvable group as part of the decomposition cascade.}, and in those cases we should expect transformers to struggle. 

One might also worry that a full theory of transformers would have to take into account the probabilistic features of the final choice of token, and how that affects iterated versions. This is something I've largely omitted, but  the overall point should generalise to probabilistic automata.\footnote{\citet{MalerA-decomposition93} gives a basic extension of the Krohn-Rhodes theorem to probabilistic automata; see \citet{DeDeoEffective11} for further discussion. } Finally, while the results I've discussed apply to FSAs, I suspect that the techniques discussed in appendix \ref{coveringappendix} are more general, and  in principle be used to analyse automata with (e.g.) bounded memory.

Of course, none of this is to yet say how efficient this process is, how generalisable it is, how learnable it is, and so on. Transformers present something of an existence proof;  there's much more to be done. The point is, rather, that if understand a transformer as giving us a representation of an $f_A$ for an automaton that can produce the corpus, there is at least a mechanism in place for something like this to happen without any need for the sort of structured representations we might use. Ultimately, this is probably a special instance of something we've known since \citet{TuringOn-computable38} introduced universal machines: emulating an automaton, even one that itself uses a lot of structure, can be done almost entirely at the level of representational content of the emulating machine.

\section{Non-deflationary conclusion: Corpus and transformations}
\label{conclusion}

In an essay on the 6th century Chinese author Liu Xie's work on literary aesthetics, \citet{OwenLiu-Xie-and-the-Discourse01} coins the evocative term `discourse machine' for the rules and requirements of classical Chinese textual production. Liu Xie's writing, says Owen, is often and clearly ``the productive rhetoric of the discourse machine processing an initial statement and amplifying it according to predictable rules.'' \citep{OwenLiu-Xie-and-the-Discourse01}. Liu Xie's own contribution is to follow in the wake of the discourse machine, subtly nudging and correcting its outputs. 

Classical Chinese writing is dense and allusive, and the training of a scholar like Liu Xie would have emphasised the expected moves in textual production. Yet this is only an extreme case. Much of what we learn as we become enculturated is how to produce text that sounds like continuations of existing text. \footnote{Including, as one ages, text that sounds like yourself. As Auden noted wistfully, ``Later in life, incidentally, [the poet] will realise how important is the art of imitation, for he will not infrequently be called upon to imitate himself.''  \citep[38]{AudenMaking62}.} We learn this, by and large, in an implicit way: by being exposed to essays or letters or novels or lyrics or emails, and then being called upon to make more of the same. This, conversely, suggests that the text we do imbibe has a kind of internal scaffolding that signals it as a type, and helps to produce more of the same. 

This is why \citet{AudenMaking62} suggests that the best way to learn to write poetry is to learn to imitate poets one admires, for it is only when one can produce imitation freely that one can start to elaborate. In a less poetic vein, there is a tremendous amount of text produced that is, by and large, slight transformations of other texts along familiar lines. Your university inbox will furnish numerous examples: an email about delayed provisioning of new microphones for hybrid classrooms ought to have a certain ponderous tone and indirect style; innovation would only raise suspicions. 

Indeed, the power of the existing corpus is so strong that it can often, as it were, make its own weather. Language can carry thought along, and sometimes we don't know where we're going until we get there. In a lovely essay on this problem, the German author Heinrich von Kleist describes it thus:  \begin{quote} \ldots when an algebraic problem arises, I look for the first preliminary statement, the equation, which expresses the given circumstances and from which later the solution can be easily deduced by calculation. But, lo and behold, if I mention it to my sister, who is sitting behind me and working, I discover facts which whole hours of brooding, perhaps, would not have revealed. Not that she literally tells them to me; for neither does she know the book of rules, nor has she studied Euler or Kastner. Nor is it that her skilful questioning leads me on to the point which matters, though this may frequently be the case. But since I always have some obscure preconception, distantly connected in some way with whatever I am looking for for, I have only to begin boldly and the mind, obliged to find an end for this beginning, transforms my confused concept as I speak into thoughts that are perfectly
clear, so that, to my surprise, the end of the sentence coincides with the desired knowledge. \citep[42]{Von-KleistOn-the-gradual51} \end{quote} The results can be far from trivial: von Kleist attributes to this  Mirabeau's rebuke of Louis XVI's envoy, and with that the start of the French Revolution.   To  add to von Kleist's perceptive description of the phenomenon,  I would only emphasise that this is only possible once we have seen enough of the kind of material we want to make in order to have learned the right moves. 

This is why it would be a misreading of the conclusions above to note that LLMs can produce text without understanding. That is true, of course, but misses the point. For  \emph{humans} arguably produce a lot of text without deep understanding. Both LLMs and humans can do so because the corpus of language already produced gives both a lot of static exemplars and also a set of permitted  \emph{transformations}: how sentences can be continued, which synonyms can be substituted in which contexts, and so on. The formal apparatus upon which the results of \citet{LiuTransformers22} involves drawing a tight link between what transformers learn and the permissible transformations of a corpus.\footnote{In the terms of appendix \ref{coveringappendix}, the transformation semigroup of an automata.} The permissible transformations need not be learned directly, however, but instead by proxy, by emulating the state transitions of an automaton that could produce the corpus.

The relationship between humans and LLMs is, in an important sense, something like convergent evolution: they perform similar tasks, but they do so with different mechanisms and different histories. Much as when we see convergent evolution in nature, we ought to look to facts about the world that explain why. Butterflies, hummingbirds, and the Australian honey possum all have long tongues because they feed on nectar from flowers. It is the shape of the flowers and the attractiveness of the nectar that primarily explains the tongue, not genetic similarity or the tongue of the last common ancestor. Similarly, I argue, transformers do not show us much about linguistic production: they are poor models of what we do. However, they show us a great deal about the \emph{corpus}, and about the role of the corpus as setting up something like a discourse machine. 

Hence we conclude with what might have seemed like a triviality at the outset: transformers learn transformations. But this is fair from a trivial, or deflationary result. The fact that an ability to learn transformations is enough to capture the corpus and produce more like it is a striking result. But it is  a striking result not about modelling language production but about the power of the corpus, however you manage to grasp it.

\section*{Acknowledgements}

Thanks to audiences at NYU, University of Pittsburgh HPS, Washington University St. Louis,  Johns Hopkins University,  CUNY Grad Center, and ANU for   feedback on a different, longer form of this paper. Thanks to Osman Attah, Andrew Barron, Ned Block, Carl Craver, David Chalmers,  Alexandre Duval, EJ Green,  Sarah-Jane Leslie, Esther Klein, Tony Licata, Sharona Lin, Mei Liou, Tania Lombrozo, Edouard Machery, Eric Mandelbaum,  Ian Phillips, and Trenton Wilson  for helpful discussions. 

This work was supported by TWCF-2020-20539 and ARC DP240100400. 

\clearpage
\newpage
\appendix
\section{Proofs}
\label{PEProof}

\subsection{Permutation Invariance} 

The following is based closely on the proof given by \citet{DufterPosition22}. Each transformer pass consists of a series of a self-attention block followed by a fully connected multilayer perceptron block, along with various norming and activation functions.  Let $P_\pi$ be a permutation matrix and $X$ an input. 

Any elementwise  function (including activation functions like $ReLU$) is  trivially permutation-invariant. Norming layers are row-wise, and any row-wise function is also permutation invariant. 

Each layer of a fully connected multilayer perceptron layer is permutation invariant. A layer is equivalent to $XW$, where $W$ is a weight matrix.  Since matrix multiplication is associative, $(P_\pi X)W = P_\pi(XW)$. 

It thus remains to be proven that the attention block is permutation invariant. Note that while one often speaks of multiple attention heads, in the forward pass these are equivalent to (and often implemented by) a single matrix multiplication, so we may treat it as such. 

Given an input $X$, the output of the attention head $M$  can be decomposed into two steps: 
\begin{align*}
 A &= XW^{(q)}(XW^{(k)})^\intercal \\
 &= X W^{(q)}W^{(k)\intercal}X^\intercal\\
M &= softmax(A)XW^{(v)}
\end{align*}
 Where $W^{(q)}, W^{(k)}$, and $W^{(v)}$ are, respectively, the query, key, and value weights. Permutation invariance holds if\mbox{}f $M(P_\pi X) = P_\pi M(X)$.
 
 Consider first the effect on $A$ of permuted input: 
 \begin{align*}
 A(P_\pi X) &= P_\pi XW^{(q)}(P_\pi XW^{(k)})^\intercal \\
 &= P_\pi X W^{(q)}W^{(k)\intercal}(P_\pi X)^\intercal\\
 &= P_\pi X W^{(q)}W^{(k)\intercal}X^\intercal P_\pi^\intercal\\
 &= P_\pi A P_\pi^\intercal
 \end{align*}
Now, $softmax(P_\pi A P_\pi^\intercal) =  P_\pi softmax(A) P_\pi^\intercal$. The left multiplication gives a row permutation, and since softmax is row-wise it doesn't matter whether you permute rows before or after. The right multiplication gives a column permutation, and since softmax only cares about the weight of an element as a function of the transformed sum of its row, not the position of the element in the row, a column-wise permutation can happen before or after softmax as well. 
 
For any permutation matrix,  $P_\pi^\intercal P_\pi = I$.  It thus follows that 
  \begin{align*}
M(P_\pi X)  &= P_\pi softmax(A) P_\pi^\intercal P_\pi XW^{(v)} \\
&= P_\pi softmax(A) XW^{(v)}
\end{align*} 
hence $M(P_\pi X) = P_\pi M(X)$  $\blacksquare$

\subsection{Four types of operation in unmasked transformers}
\label{fourtypes}

I have discussed permutation invariance as a function of the whole transformer. However, as noted in the main text, permutation invariance comes about because each block is itself permutation invariant. We may see this by grouping the operations  performed in unmasked transformers into four types.



The residual stream can be seen as a set of $d$-dimensional vectors, and each type of operation is an operation on those vectors. In addition the operation of adding something back to the residual stream, there are four basic kinds of operation that can be done on such a set: \begin{enumerate}
\item Each vector can be independently \emph{rescaled} ($softmax$ and $layernorm$) 
\item Each vector can be independently  \emph{weighted} by a second matrix. (the matrix multiplications involved in the key, value, and query matrices, and the MLP block) 
\item Each vector can be \emph{compared} to one another in an all-to-all fashion (The derivation of the attention matrix) 
\item We can derive a \emph{weighted sum} of every vector, with the weights given by a second matrix. (Multiplication of the attention matrix by the input to the block)  
\end{enumerate}
 The important thing to emphasise at this point is that each operation is either completely row-wise (rescaling, weighting) or involves \emph{every} vector (comparing, weighted sum). There is no conditional comparison, and no privileged partnerships between one vector and another. 

Permutation invariance can thus be seen as a consequence of the fact that the basic operations are all themselves permutation-invariant. This is also why, as emphasised in the main text, substring invariance of masked transformers still implies permutation invariance of the preceding tokens: masking only restricts the attention matrix, but the operations performed are identical to that of basic transformers, and hence themselves permutation invariant.

\subsection{Substring Invariance}
\label{substringproof}

The masking step makes the attention matrix lower triangular.  The effect depends on whether the mask is applied before or after the softmax operation. If it were applied after, substring invariance follows straightforwardly from the properties of matrix multiplication making the attention mask upper triangular means that an added row cannot affect the subsequent operations except for that row itself.  (Adding a row would change a $t\times t$ attention matrix to a  $(t+1) \times (t+1)$ matrix, the last column of which is all zeros except for the $t+1$th row.) In general, masking in this manner means that the effect of the attention block only takes into account information in a token and those immediately preceding. Since all subsequent operations are row-wise, that property will be preserved. 

If softmax is applied after the masking, however (which appears to be more common version,  and is conceptually cleaner)  then we get only approximate substring invariance. The reason is that the classic softmax function applied to an element $r_i$ of a row: 
  \begin{align*}
\sigma(r_i) = \frac{e^{r_{i}}}{\sum_{j=1}^t e^{r_{j}}} 
\end{align*} 
is sensitive to the number of zeros in the row: larger rows will have a larger denominator. Each of the zeroed entries in the upper triangular will have the lowest value in the row, and the larger the string the lower the value. This value will still be nonzero (and differ from row to row), which means that adding a row will have a very small effect on subsequent calculations. The additional effect of new rows gets smaller as inputs get longer.

This effect is detectable, especially at short string lengths, but it is very small. Again, as the quote from \citet{VaswaniAttention17} in the main text emphasises, the \emph{point} of masking is to keep earlier tokens from effecting later ones. Insofar as there is an effect, it is not specific to the content of any later tokens: it is better conceptualised as a very slight overall effect of the input length. This is, I suggest, entirely consistent with linearity (many linear data structures have access to their length), and so should not  affect the main argument in the text.

\section{More on positional encoding}
\label{positionalencoding}

Positional encoding can in fact be done in many ways, each mathematically equivalent but differing in other important training-relevant properties. I have focused on the simplest strategy, used by  \citep{VaswaniAttention17}, but  more complex strategies can be seen as differing implementations of it  \citep{DufterPosition22}. The choice of position matrix itself is also relatively unimportant for this argument, so long as it satisfies certain intuitive constraints (monotonicity, translation invariance, and symmetry \citealp{WangOn-position20}). Informally, you want your position vectors to behave like distances. There also need to be relatively few collisions ---that is, there are relatively few token vectors $w,v$ and position vectors $a,b$ s.t.\ $w+a \approx v+b$.  While one can find collisions, this condition is (unsurprisingly) mostly met.

The fact that position encoding can be done by simply adding vectors is usually presented quickly, but it is worth lingering on.\footnote{Thanks to Mei Liou for encouraging me to linger.}  \emph{Prima facie}, we have a set of vectors carefully encoding the tokens of the input. Then we take an unrelated set of vectors and mash them together. Why does this work?   The key, I think,  lies in the mathematical properties of the dot product. Most of what happens in a transformer is repeated matrix multiplications between the residual stream and sets of weights (what I've called `reweighting').  Matrix multiplication can be viewed as a bunch of repeated {dot products}  between vectors: if you have $AB = C$, then $C_{i,j}$ is the the dot product of the $i$th row of $A$ and the $j$th column of $B$. 

This also gives a handy alternative interpretation of reweighting: since the dot product is proportional to the cosine similarity between two vectors, the result is a parallel similarity comparison between each position in the residual stream and a set of vectors.The MLP layers in particular can be treated as a kind of fuzzy key-value memory  \citep{GevaTransformer20}. The dot product has the handy property of being distributive over vector addition: that is, ${w} \cdot (u+v) = ({w} \cdot {u}) + ({w} \cdot {v})$.  So when we input $X=(E+P)$, the next operations we perform will, in some important sense, be applied both to the token vector and the associated position vector. The results are summed, which means that subsequent comparisons to \emph{that} vector can be seen as the sum of two comparisons, and so on. 

I noted in section \ref{perminvariance} that positional encoding does not affect permutation invariance. Two small technical objections are worth dispatching. First, the encoding and decoding phases of an LLM---i.e. the move from words to matrices and back from matrices to words---\emph{does} care about ordering, in the sense that the first word becomes the first row of $X$, the second the second, and so on. Tokens are extracted back out in the same order. However, the fact that something is in the first row of the input (or the residual stream or whatever) doesn't introduce additional structure, because that fact doesn't play any role in the intermediate processing. Think of each token in the input as getting a tag corresponding to its place in the matrix.  This tag persists across transformations of the information corresponding to that token. The tag only persists because it is never altered, never read, and plays no role whatsoever in the subsequent computation. Any processing that involves positional information must do so on the embedded position information, not on the tag.   Second, and relatedly, the results of the attention and MLP blocks are added back to the residual stream.  This requires the block output and the residual stream to align on rows. But that doesn't mean that the row ordering is  important: it just means that you're performing an operation on each element of the stream and adding the result back to that very element. The fact that this is the (e.g.) the first element calculated rather than the third can't make a difference. 

Section \ref{basictransformersformat} claimed that positional information is not generally preserved.  Positional information is present in the first transformer layer, and we can show this by looking at the cosine similarity between the residual stream and the position matrix. Figure \ref{posfig} shows the result of doing so on the same sample input used in figure \ref{AllLogits}.  Columns correspond to the residual stream after each block, rows to  positions. Note that position information is reliably extracted in column 0 (corresponding to the original input), and then  disappears rapidly after the first layer.

\begin{figure}[h]
\begin{center}
\begin{tabular}{l | r r r r r r r r r r r r r}
& \multicolumn{13}{c}{After hidden layer} \\
& 0 & 1 & 2 & 3 & 4 & 5 & 6 & 7 & 8 & 9 & 10 & 11 & 12 \\
\hline
0  &  0 & 0 & 0 & 5 & 5 & 0 & 2 & 6 & 2 & 2 & 2 & 2 & 5 \\
1  &  1 & 1 & 5 & 5 & 6 & 5 & 6 & 6 & 6 & 6 & 1 & 1 & 5 \\
2  &  2 & 1 & 6 & 6 & 6 & 6 & 6 & 6 & 6 & 6 & 6 & 1 & 5 \\
3  &  3 & 7 & 7 & 7 & 7 & 7 & 7 & 7 & 7 & 7 & 6 & 5 & 7 \\
4  &  4 & 5 & 5 & 5 & 5 & 5 & 5 & 5 & 5 & 5 & 5 & 4 & 5 \\
5  &  5 & 6 & 5 & 5 & 5 & 5 & 5 & 5 & 5 & 5 & 4 & 1 & 5 \\
6  &  6 & 7 & 7 & 7 & 7 & 7 & 7 & 7 & 7 & 4 & 1 & 1 & 6 \\
7  &  7 & 7 & 7 & 7 & 7 & 8 & 8 & 8 & 8 & 4 & 4 & 4 & 8 \\
8  &  8 & 7 & 7 & 7 & 7 & 7 & 7 & 6 & 7 & 6 & 5 & 4 & 0 \\
\end{tabular}
\end{center}
\caption{Cosine similarity between each row of the residual stream and the position matrix after the operation of a full block in the OpenAIGPT model.  Highest-probability match  for each row shown. Comparison has been restricted to the first 9 positions of the position matrix. Column 0 is the original input with positional encoding.  Note how rapidly position information is lost as you move left to right. The figure is based on a sample sentence (``when lions eat humans they get a bad'') fed to an instance of OpenAIGPT. }
\label{posfig}
\end{figure}

\section{Automata, Covering, and the Krohn-Rhodes theorem}
\label{coveringappendix}

The story told in section \ref{positivestory} moves quickly through a number of details. Here is a bit of an elaboration and gesture at the bigger ideas in the background. I rely in particular on the presentation of the Krohn-Rhodes theorem found in \citep{EilenbergAutomata74b}.  

A finite state automata (FSA) is defined by a tuple $X = (\Sigma, Q, \delta)$, where $\Sigma$ is a finite input alphabet of symbols , $Q$ a finite set of states, and $\delta$  a transition function $\delta:  \Sigma \times Q \rightarrow Q$. At each computational step, an FSA starts in some state $q\in Q$, reads in a new symbol $\sigma \in \Sigma$, and transitions to a new state $q^\prime = \delta(\sigma, q)$. 

Say that an FSA $Y$ \emph{covers} $X$ just in case $X$ and $Y$ accept the same input alphabet, and there is surjective partial function $\phi: Q_Y \rightarrow Q_X$ such that, for any input $\sigma\in\Sigma$, $\phi(\delta_Y(\sigma, q_Y)) = \delta_X(\sigma, \phi(q_Y))$. Covering is a way to make the notion of `emulation' precise. Intuitively, to say that $Y$ covers $X$ is to say that we can map from the states of $Y$ to $X$ in a way that lets us say $Y$ is doing the same thing as $X$, though possibly with more states involved. 

Now, if we think of the partial application of $\delta$ for any input symbol we find that it specifies a transformation function $f_\sigma: Q\rightarrow Q$. Each transformation---or, if you'd like, the characteristics of a single row in the machine table---falls into one of three possibilities. First, a symbol may be a \emph{reset}, taking any state into one specific one or else leaving it unchanged. Second, a symbol may be a \emph{permutation} of the states; this is true if the row of the machine table contains all of the states in $Q$.\footnote{Note that identity rows count as both permutations and resets on this definition. This is done deliberately, to allow for the possibility of transformation monoids of both types.} Intuitively a permutation row is one where the action of the symbol is entirely dependent on the current state, while a reset row is one where the action of the symbol is entirely indifferent to the current state. Finally, a row might be a \emph{mixed} row, which is just one that is neither a reset nor a permutation. 

Now, let the set $T$ be the closure under function composition of each transformation function. Every input sequence will specify a transformation in $T$. $T$ itself gives a \emph{transformation semigroup} of $X$, with $X_Q$ being the underlying set . The move from an automaton to its transformation semigroup allows us to consider automata from an algebraic point of view. This allows us to bring the considerable resources of abstract algebra to bear on questions about automata. We must use transformation semigroups (rather than transformation groups) because many rows of a machine table give rise to non-invertible transformation functions.\footnote{In practice, many of the proofs mentioned will work with transformation monoids---i.e. transformation semigroups with an additional identity element added---because this simplifies the resulting proofs.} 

The task of finding an automata to cover a given $A$ has an equivalent algebraic formulation: we can cover $A$ just in case we can find an automata $A^\prime$ with a transformation semigroup of which the transformation semigroup of $A$ is a subsemigroup. The Krohn-Rhodes theorem (in simplified form) implies\footnote{Originally presented in \citep{KrohnAlgebraic65}. There have been a number of more streamlined proofs since; I have relied especially on the presentation in \citep{EilenbergAutomata74b}.}  that every transformation semigroup can be decomposed into the wreath product of reset and permutation semigroups. There is usually not a unique such decomposition, and the sense of `decomposition' involved is not isomorphism between the wreath product and the target but rather the existence of a surjective partial function from the wreath product to the target. 

Returning to automata, the Krohn-Rhodes theorem has the consequence that any automaton $X$ can be covered by a special `cascade' of simpler automata. A cascade of machines is one where there is an ordering on automata in a sequence, and the machine table of each automaton in the sequence takes into account the current input symbol and the states of the previous machines in the cascade \citep{ZeigerCascade67,ZeigerYet-another67}. A cascade is feedforward in the sense that information flows only down the sequence; the only loop is the implied return to the beginning of the sequence after the input symbol is consumed. (Note here the similarity with information flow in transformers!) Each  element in the cascade corresponds to an element in the wreath product that decomposes the transformation semigroup of $X$. 

Now, say that an automaton is a reset automaton if every $t\in T$ of its corresponding transformation semigroup is a reset (or the identity), and a permutation automaton if every $t\in T$ is a permutation (or, equivalently, if $T$ is a group). Hence each element of the cascade, by the Krohn-Rhodes theorem, either a permutation or a reset automaton.

In an intuitive sense, the Krohn-Rhodes theorem says that we can, in effect, factor out the actions of an automaton into bits that correspond to groups (pure permutation-automata) and bits that lose information, because all they care about is the input (pure reset-automata). As \citeauthor{DeDeoEffective11}  puts it:\begin{quote}
One of the interesting features of the Krohn-Rhodes theorem is the difference in treatment of the reversible (group symmetry) and irreversible transformations of the system. While the groups resolve themselves into a non-trivial catalog of simple subunits, there are no `irreducible' semigroups
of dissipation beyond the flip-flop. There are distinct and irreducible groups of reversible computations, but the irreversible aspects of a computation decompose finally into collections of pure identity-resets.\citep[4-5]{DeDeoEffective11}  \end{quote}
Another way of thinking about this, which is closer to the presentation of  \citep{ZeigerCascade67}, is to think of the cascade of simple automata as like delay lines. A cascade of automata, then, can be seen as showing how much of the context of previous inputs needs to be incorporated in order to understand the action of an automaton on its current input. 

While relatively obscure now, the Krohn-Rhodes theorem (and algebraic approaches to computation more generally)  used to be a hot topic in automata theory.  Several advanced textbooks from the late 1960s emphasise the algebraic understanding of machines.\footnote{See e.g. \citep{GinzburgAlgebraic68,KalmanTopics69, EilenbergAutomata74a,EilenbergAutomata74b}.}  The general approach fell out of fashion,\footnote{As Mahler notes of Krohn and Rhodes:``For some time in the 60s and 70s, their theorem, which got them 2 simultaneous PhD titles from Harvard and MIT, respectively, was considered to be a cornerstone of automata theory. When I started to look at the topic in the late 80s the results have been practically forgotten in the Computer Science mainstream, excluding some specialized islands'' \citeyear[260]{MalerOn-the-Krohn-Rhodes10}.}  but there has been something of a resurgence of interest in the Krohn-Rhodes theorem in recent years. This is partly due to the potential application to transformers, though several authors have also  suggested that this approach gives a more general way to think about hierarchical understanding of automata and complex systems \citep{Egri-NagyHierarchical08,DeDeoEffective11}. 

Returning to transformers, and in particular the model of \citet{LiuTransformers22}, transformers have two useful properties. First, one can show that they are---architecturally speaking---capable of calculating the transformation semigroup of an automata cascade in a relatively straightforward way, and can very often do so in a parallel manner that gives efficient shortcut solutions.  Roughly speaking, the attention blocks handle resets, while the MLP blocks handle the group-like permutations.  Second, they provide evidence that transformers trained in the ordinary way can learn actions corresponding to the Krohn-Rhodes decomposition of automata chosen to generate training data. The implication, then, is that transformers trained on a corpus of natural language learn an efficient calculation of the transformation semigroup  of an automata that could give rise to the training corpus in the first place. 

Since shortcut solutions are often bound to input length, this also explains why and when transformers struggle with inputs longer than their training set. Finally, and intriguingly,  \citeauthor{LiuTransformers22} suggest that there ought to be an important class of failures that arise when some of the automata in the decomposition are  non-solvable groups, since there are not efficient shortcuts for emulating non-solvable groups. 

\citet{LiuTransformers22}'s work is in many ways speculative, and I have gone quickly over the details. However, the takeaway points I think are threefold: First, that the action of any automaton corresponds to a transformation semigroup acting on the set of states. Second, that that transformation semigroup can be covered by specific algebraic products of much simpler semigroups. And, third, that this simpler covering, while of a much larger size, often admits of parallel `shortcut' calculations of the sort that transformers appear to be able to implement via all-to-all comparisons of tokens.


\newpage
\clearpage
\nocite{MilliereA-Philosophical24-1}

\begin{thebibliography}{}

\bibitem[Abelson et~al., 1996]{AbelsonStructure96}
Abelson, H., Sussman, G.~J., and Sussman, J. (1996).
\newblock {\em Structure and interpretation of computer programs}.
\newblock MIT Press, Cambridge.

\bibitem[Alammar, 2018]{AlammarThe-Illustrated18}
Alammar, J. (2018).
\newblock The illustrated transformer.
\newblock https://jalammar.github.io/illustrated-transformer/.

\bibitem[Auden, 1962]{AudenMaking62}
Auden, W. (1962).
\newblock Making, knowing and judging.
\newblock In {\em The Dyer's Hand, and Other Essays}, pages 31--60. Random
  House.

\bibitem[Bender et~al., 2021]{BenderOn-the-dangers21}
Bender, E.~M., Gebru, T., McMillan-Major, A., and Shmitchell, S. (2021).
\newblock On the dangers of stochastic parrots: Can language models be too big?
\newblock In {\em Proceedings of the 2021 ACM conference on fairness,
  accountability, and transparency}, pages 610--623.

\bibitem[Butlin et~al., 2023]{ButlinConsciousness23}
Butlin, P., Long, R., Elmoznino, E., Bengio, Y., Birch, J., Constant, A.,
  Deane, G., Fleming, S.~M., Frith, C., Ji, X., et~al. (2023).
\newblock Consciousness in artificial intelligence: {I}nsights from the science
  of consciousness.
\newblock {\em arXiv preprint arXiv:2308.08708}.

\bibitem[Chalmers, 2023a]{ChalmersDoes23}
Chalmers, D. (2023a).
\newblock Does thinking require sensory grounding? {F}rom pure thinkers to
  large language models.

\bibitem[Chalmers, 2023b]{ChalmersCould23}
Chalmers, D.~J. (2023b).
\newblock Could a large language model be conscious?
\newblock {\em arXiv preprint arXiv:2303.07103}.

\bibitem[Chiang, 2023]{ChiangChatGPT23}
Chiang, T. (2023).
\newblock {ChatGPT} is a blurry {JPEG} of the web.
\newblock \emph{The New Yorker} Feb 9.

\bibitem[Chomsky, 1956]{ChomskyThree56}
Chomsky, N. (1956).
\newblock Three models for the description of language.
\newblock {\em IRE Transactions on information theory}, 2(3):113--124.

\bibitem[Chomsky, 1959]{ChomskyOn-certain59}
Chomsky, N. (1959).
\newblock On certain formal properties of grammars.
\newblock {\em Information and control}, 2(2):137--167.

\bibitem[Cordonnier et~al., 2019]{CordonnierOn-the-relationship19}
Cordonnier, J.-B., Loukas, A., and Jaggi, M. (2019).
\newblock On the relationship between self-attention and convolutional layers.
\newblock {\em arXiv preprint arXiv:1911.03584}.

\bibitem[DeDeo, 2011]{DeDeoEffective11}
DeDeo, S. (2011).
\newblock Effective theories for circuits and automata.
\newblock {\em Chaos: An Interdisciplinary Journal of Nonlinear Science},
  21(3).

\bibitem[Devlin, 2018]{DevlinBert:18}
Devlin, J. (2018).
\newblock {BERT}: Pre-training of deep bidirectional transformers for language
  understanding.
\newblock {\em arXiv preprint arXiv:1810.04805}.

\bibitem[Dufter et~al., 2022]{DufterPosition22}
Dufter, P., Schmitt, M., and Sch{\"u}tze, H. (2022).
\newblock Position information in transformers: An overview.
\newblock {\em Computational Linguistics}, 48(3):733--763.

\bibitem[Egri-Nagy and Nehaniv, 2008]{Egri-NagyHierarchical08}
Egri-Nagy, A. and Nehaniv, C.~L. (2008).
\newblock Hierarchical coordinate systems for understanding complexity and its
  evolution, with applications to genetic regulatory networks.
\newblock {\em Artificial Life}, 14(3):299--312.

\bibitem[Eilenberg, 1974a]{EilenbergAutomata74a}
Eilenberg, S. (1974a).
\newblock {\em Automata, languages, and machines, Volume A}.
\newblock Academic press, New York.

\bibitem[Eilenberg, 1974b]{EilenbergAutomata74b}
Eilenberg, S. (1974b).
\newblock {\em Automata, languages, and machines: Volume B}.
\newblock Academic press, New York.

\bibitem[Fodor, 1975]{JFLOT}
Fodor, J. (1975).
\newblock {\em The Language of Thought}.
\newblock Thomas Y.\ Crowell Company, New York.

\bibitem[Geva et~al., 2020]{GevaTransformer20}
Geva, M., Schuster, R., Berant, J., and Levy, O. (2020).
\newblock Transformer feed-forward layers are key-value memories.
\newblock {\em arXiv preprint arXiv:2012.14913}.

\bibitem[Ginzburg, 1968]{GinzburgAlgebraic68}
Ginzburg, A. (1968).
\newblock {\em Algebraic theory of automata}.
\newblock Academic Press, New York.

\bibitem[Guttag, 1977]{GuttagAbstract77}
Guttag, J. (1977).
\newblock Abstract data types and the development of data structures.
\newblock {\em Communications of the ACM}, 20(6):396--404.

\bibitem[Hagberg et~al., 2008]{HagbergExploring08}
Hagberg, A., Swart, P.~J., and Schult, D.~A. (2008).
\newblock Exploring network structure, dynamics, and function using
  \mbox{NetworkX}.
\newblock Technical report, Los Alamos National Laboratory (LANL), Los Alamos,
  NM (United States).

\bibitem[Haugeland, 1998]{HaugelandRepresentational98}
Haugeland, J. (1998).
\newblock Representational genera.
\newblock In {\em Having Thought: Essays in the Metaphysics of Mind}, pages
  171--206. Harvard University Press, Cambridge MA.

\bibitem[Kalman et~al., 1969]{KalmanTopics69}
Kalman, R.~E., Falb, P.~L., and Arbib, M.~A. (1969).
\newblock {\em Topics in mathematical system theory}.
\newblock McGraw-Hill, New York.

\bibitem[Klein, 2025]{KleinComputational26}
Klein, C. (2025).
\newblock Computational individuation: Isomorphism, not indeterminacy.
\newblock {\em Analysis (forthcoming)}, https://doi.org/10.1093/analys/anaf003.

\bibitem[Klein and Barron, 2024]{KleinComparing24}
Klein, C. and Barron, A.~B. (2024).
\newblock Comparing cognition across major transitions using the hierarchy of
  formal automata.
\newblock {\em Wiley Interdisciplinary Reviews: Cognitive Science}, page e1680.

\bibitem[Klein and Clutton, 2021]{KleinWhat21}
Klein, C. and Clutton, P. (2021).
\newblock What is the job of the job description challenge? {A} study in
  esoteric and exoteric semantics.
\newblock In Calzavarini, F. and Viola, M., editors, {\em Neural Mechanisms},
  pages 449--463. Springer.

\bibitem[Knuth, 1997]{KnuthThe-Art-of-Computer97}
Knuth, D.~E. (1997).
\newblock {\em The Art of Computer Programming Volume 1: {F}undamental
  Algorithms}, volume~1.
\newblock Addison--Wesley, Reading, Massachusetts, 3rd edition.

\bibitem[Krohn and Rhodes, 1965]{KrohnAlgebraic65}
Krohn, K. and Rhodes, J. (1965).
\newblock Algebraic theory of machines. i. prime decomposition theorem for
  finite semigroups and machines.
\newblock {\em Transactions of the American Mathematical Society},
  116:450--464.

\bibitem[Liu et~al., 2022]{LiuTransformers22}
Liu, B., Ash, J.~T., Goel, S., Krishnamurthy, A., and Zhang, C. (2022).
\newblock Transformers learn shortcuts to automata.
\newblock {\em arXiv preprint arXiv:2210.10749}.

\bibitem[Maler, 1993]{MalerA-decomposition93}
Maler, O. (1993).
\newblock A decomposition theorem for probabilistic transition systems.
\newblock In {\em STACS 93: 10th Annual Symposium on Theoretical Ascpects of
  Computer Science W{\"u}rzburg, Germany, February 25--27, 1993 Proceedings
  10}, pages 323--332. Springer.

\bibitem[Maler, 2010]{MalerOn-the-Krohn-Rhodes10}
Maler, O. (2010).
\newblock On the {K}rohn-{R}hodes cascaded decomposition theorem.
\newblock In {\em Time for Verification: Essays in Memory of Amir Pnueli},
  pages 260--278. Springer.

\bibitem[Marr, 1982]{MarrVision:82}
Marr, D. (1982).
\newblock {\em Vision: A computational investigation into the human
  representation and processing of visual information}.
\newblock WH Freeman, New York.

\bibitem[McCoy et~al., 2024]{McCoyEmbers24}
McCoy, R.~T., Yao, S., Friedman, D., Hardy, M.~D., and Griffiths, T.~L. (2024).
\newblock Embers of autoregression show how large language models are shaped by
  the problem they are trained to solve.
\newblock {\em Proceedings of the National Academy of Sciences},
  121(41):e2322420121.

\bibitem[Milli{\`e}re and Buckner, 2024a]{MilliereA-Philosophical24-1}
Milli{\`e}re, R. and Buckner, C. (2024a).
\newblock A philosophical introduction to language models--part {I}:
  {C}ontinuity with classic debates.
\newblock {\em arXiv preprint arXiv:2401.03910}.

\bibitem[Milli{\`e}re and Buckner, 2024b]{MilliereA-Philosophical24-2}
Milli{\`e}re, R. and Buckner, C. (2024b).
\newblock A philosophical introduction to language models-part {II}: {T}he way
  forward.
\newblock {\em arXiv preprint arXiv:2405.03207}.

\bibitem[Mollo and Vernazzani, 2023]{MolloThe-formats23}
Mollo, D.~C. and Vernazzani, A. (2023).
\newblock The formats of cognitive representation: A computational account.
\newblock {\em Philosophy of Science}, pages 1--20.

\bibitem[Neisser, 2014]{NeisserCognitive67}
Neisser, U. (1967/2014).
\newblock {\em Cognitive psychology: Classic edition}.
\newblock Psychology press.

\bibitem[nostalgebraist, 2020]{nostalgebraistinterpreting20}
nostalgebraist (2020).
\newblock interpreting gpt: the logit lens.
\newblock
  https://www.lesswrong.com/posts/AcKRB8wDpdaN6v6ru/interpreting-gpt-the-logit-lens.

\bibitem[Olah et~al., 2020]{OlahZoom20}
Olah, C., Cammarata, N., Schubert, L., Goh, G., Petrov, M., and Carter, S.
  (2020).
\newblock Zoom in: An introduction to circuits.
\newblock {\em Distill}.
\newblock https://distill.pub/2020/circuits/zoom-in.

\bibitem[Owen, 2001]{OwenLiu-Xie-and-the-Discourse01}
Owen, S. (2001).
\newblock {L}iu {X}ie and the discourse machine.
\newblock In {\em A Chinese Literary Mind: Culture, Creativity, and Rhetoric in
  Wenxin Diaolong}, pages 175--192. Stanford University Press, Stanford.

\bibitem[Pavlick, 2023]{PavlickSymbols23}
Pavlick, E. (2023).
\newblock Symbols and grounding in large language models.
\newblock {\em Philosophical Transactions of the Royal Society A},
  381(2251):20220041.

\bibitem[Pylyshyn, 2003]{PylyshynReturn03}
Pylyshyn, Z. (2003).
\newblock Return of the mental image: {A}re there really pictures in the brain?
\newblock {\em Trends in cognitive sciences}, 7(3):113--118.

\bibitem[Pylyshyn, 1984]{PylyshynComputation84}
Pylyshyn, Z.~W. (1984).
\newblock {\em Computation and cognition}.
\newblock Cambridge University Press, Cambridge.

\bibitem[Quilty-Dunn et~al., 2023]{Quilty-DunnThe-best23}
Quilty-Dunn, J., Porot, N., and Mandelbaum, E. (2023).
\newblock The best game in town: The reemergence of the language-of-thought
  hypothesis across the cognitive sciences.
\newblock {\em Behavioral and Brain Sciences}, 46:e261.

\bibitem[Shepard and Metzler, 1971]{ShepardMental71}
Shepard, R.~N. and Metzler, J. (1971).
\newblock Mental rotation of three-dimensional objects.
\newblock {\em Science}, 171(3972):701--703.

\bibitem[Smith, 1996]{Cantwell-SmithOn-the-origin96}
Smith, B.~C. (1996).
\newblock {\em On the origin of objects}.
\newblock The MIT Press, Cambridge.

\bibitem[Sterling and Laughlin, 2015]{SterlingPrinciples15}
Sterling, P. and Laughlin, S. (2015).
\newblock {\em Principles of neural design}.
\newblock MIT Press, Cambridge.

\bibitem[Strachan et~al., 2024]{StrachanTesting24}
Strachan, J.~W., Albergo, D., Borghini, G., Pansardi, O., Scaliti, E., Gupta,
  S., Saxena, K., Rufo, A., Panzeri, S., Manzi, G., et~al. (2024).
\newblock Testing theory of mind in large language models and humans.
\newblock {\em Nature Human Behaviour}, pages 1--11.

\bibitem[Turing, 1938]{TuringOn-computable38}
Turing, A.~M. (1938).
\newblock On computable numbers, with an application to the
  entscheidungsproblem. a correction.
\newblock {\em Proceedings of the London Mathematical Society}, 2(1):544.

\bibitem[Vaswani et~al., 2017]{VaswaniAttention17}
Vaswani, A., Shazeer, N., Parmar, N., Uszkoreit, J., Jones, L., Gomez, A.~N.,
  Kaiser, {\L}., and Polosukhin, I. (2017).
\newblock Attention is all you need.
\newblock {\em Advances in neural information processing systems},
  30:6000--6010.

\bibitem[Von~Kleist and Hamburger, 1951]{Von-KleistOn-the-gradual51}
Von~Kleist, H. and Hamburger, M. (1951).
\newblock On the gradual construction of thoughts during speech.
\newblock {\em German Life and Letters}, 5(1):42--46.

\bibitem[Wang et~al., 2020]{WangOn-position20}
Wang, B., Shang, L., Lioma, C., Jiang, X., Yang, H., Liu, Q., and Simonsen,
  J.~G. (2020).
\newblock On position embeddings in bert.
\newblock In {\em International Conference on Learning Representations}.

\bibitem[Wirth, 1976]{WirthAlgorithms76}
Wirth, N. (1976).
\newblock {\em Algorithms+ Data Structures= Programs}.
\newblock Prentice Hall.

\bibitem[Xu et~al., 2024]{XuPermutation24}
Xu, H., Xiang, L., Ye, H., Yao, D., Chu, P., and Li, B. (2024).
\newblock Permutation equivariance of transformers and its applications.
\newblock In {\em Proceedings of the IEEE/CVF Conference on Computer Vision and
  Pattern Recognition}, pages 5987--5996.

\bibitem[Yildirim and Paul, 2024]{YildirimFrom24}
Yildirim, I. and Paul, L. (2024).
\newblock From task structures to world models: {W}hat do {LLMs} know?
\newblock {\em Trends in Cognitive Sciences}.

\bibitem[Zeiger, 1967a]{ZeigerCascade67}
Zeiger, H.~P. (1967a).
\newblock Cascade synthesis of finite-state machines.
\newblock {\em Information and Control}, 10(4):419--433.

\bibitem[Zeiger, 1967b]{ZeigerYet-another67}
Zeiger, P. (1967b).
\newblock Yet another proof of the cascade decomposition theorem for finite
  automata.
\newblock {\em Mathematical systems theory}, 1(3):225--228.

\end{thebibliography}

\end{document}